\documentclass{article}

\usepackage{PRIMEarxiv}

\usepackage[utf8]{inputenc} 
\usepackage[T1]{fontenc}    
\usepackage{hyperref}       
\usepackage{url}            
\usepackage{booktabs}       
\usepackage{amsfonts}       
\usepackage{nicefrac}       
\usepackage{microtype}      
\usepackage{lipsum}
\usepackage{fancyhdr}       
\usepackage{graphicx}       
\graphicspath{{media/}}     

\title{Happy or grumpy? A Machine Learning Approach to Analyze the Sentiment of Airline Passengers' Tweets}

\author{
  Shengyang Wu \\
  Department of Computer Science \\
  Purdue University \\
  West Lafayette, IN, 47907\\
  \texttt{wu1632@purdue.edu} \\
   \And
  Yi Gao, Ph.D. \\
  School of Aviation and Transportation Technology \\
  Purdue University \\
  West Lafayette IN, 47907\\
  \texttt{yigao@purdue.edu} \\
}

\begin{document}
\maketitle

\begin{abstract}

As one of the most extensive social networking services, Twitter has more than 300 million active users as of 2022. Among its many functions, Twitter is now one of the go-to platforms for consumers to share their opinions about products or experiences, including flight services provided by commercial airlines. This study aims to measure customer satisfaction by analyzing  sentiments of Tweets that mention airlines using a machine learning approach. Relevant Tweets are retrieved from Twitter's API and  processed through tokenization and vectorization. After that, these processed vectors are passed into a pre-trained machine learning classifier to predict the sentiments. In addition to sentiment analysis, we also perform lexical analysis on the collected Tweets to model keywords' frequencies, which provide meaningful contexts to facilitate the interpretation of sentiments. We then apply time series methods such as Bollinger Bands to detect abnormalities in sentiment data. Using historical records from January to July 2022, our approach is proven to be capable of capturing sudden and significant changes in passengers' sentiment. This study has the potential to be developed into an application that can help airlines, along with several other customer-facing businesses, efficiently detect abrupt changes in customers' sentiments and take adequate measures to counteract them.   
\end{abstract}


\section{Introduction}
Airlines are a customer-facing industry. Knowing how passengers perceive airline travel products and services is crucial for these multi-billion dollar companies to improve their operations. Traditionally, most companies, including airlines, relied on expensive and slow market surveys to collect customer feedback \cite{Jsiwu}. Nowadays, with the increasing popularity of Social Network Services (SNS) applications, efficiently collecting genuine and unsolicited opinions has become possible and more accessible than ever before.  

Among many different SNS applications, Twitter is a popular platform for people to openly express their opinions about public matters or commercial products. The open nature of Twitter has unlocked a door for potential research into valuable and abundant User-Generated Content (UGC). In the airline realm, there have already been studies looking into this direction. We review and summarize such studies in the literature review section. A common shortcoming of these studies is that they collect historical Tweets all at once and do not monitor the change in customers' sentiment over time. This has become the motivation of this study. 

In this study, we aim to build a pipeline that can periodically retrieve Tweets about selected airlines, analyze passengers' collective sentiments toward an airline, monitor the ups and downs of such sentiments on a daily basis, detect extreme sentiment breaks, and substantiate these breaks with actual events. By automating this process, we wish to provide airlines and other customer-facing industries with a validated approach to continuously monitor customers' perceptions and opinions. This approach can also detect service disruptions and other unexpected events early so  airlines can quickly respond to such emergencies.  

The rest of the article is organized as follows: Section 1.2 reviews previous work on the sentiment analysis of Tweets and the gap missed by these studies. Section 2 explains in detail the methods that we use to perform research. Section 3 presents the results of our study, including time series plots of the overall sentiment and word frequency analysis. Finally, the conclusions and implications of the study are synthesized in Section 4, along with possible future works.

\subsection{Literature Review}
Airlines have been collecting customer feedback and trying to understand customer satisfaction to improve servivce \cite{BTS1}. Measuring customer feedback was a difficult task before the Internet era. J.D. Power, a global leader in consumer insights, advisory services, and data and analytics, started a rush of surveys to collect consumer feedback back in 1968. It took quite a few years until a reliable source of information regarding customer feedback could be established \cite{Jsiwu}. Nowadays, companies can take advantage of UGC with regard to their products and services shared by customers in SNS applications. Sentiment/topic analysis of UGC has the potential to provide valuable insights for many businesses, including the airline industry. 

Twitter sentiment analysis handles the problem of analyzing Tweets and gauging the sentiments they express. Unlike traditional sentiment analysis tasks, Twitter sentiment analysis has some unique challenges. From a survey presented by Giachanou and Crestani, Twitter sentiment analysis is a non-trivial task. It differs considerably from sentiment analysis on conventional text in aspects such as text length, topic relevance, incorrect English, data sparsity, etc. \cite{litrev1}. Currently, there are four different approaches in Twitter sentiment analysis: Machine Learning, Lexicon-Based, Hybrid, and Graph-Based. In the machine learning approach, which is what we use in this study, a machine-learning method is employed, and a classifier is built using a number of different features. Some of the most applied Twitter sentiment analysis classifiers include Naive Bayes (NB), Maximum Entropy (MaxEnt), Support Vector Machines (SVM), Multinomial Naive Bayes (MNB), Logistic Regression (LR), Random Forest (RF), and Conditional Random Field (CRF), each with its  advantage \cite{litrev1}.

There have been works done on sentiment analysis for the airline industry. Rane and Kumar evaluated different classifiers' performances on a pre-labeled airline Tweets dataset from Kaggle. Based on their evaluation, Adaptive Boost (AdaBoost), RF, and SVM algorithms yield the top three best prediction accuracy scores \cite{litrev2}. Additionally, Kumar, Sheshanna, and Babu proposed a more advanced technique with artificial neural networks (ANN) using the same dataset from Kaggle and compared the results with classifiers such as SVM and Decision Tree (DT). They conclude that ANN performed better than DT and slightly better than SVM when comparing their accuracy scores, precision scores, recalls, and F1-Scores \cite{litrev3}.

There are also many other similar studies conducted on the sentiment analysis of airline-related Tweets. Although each proposed different analysis techniques, they only performed Twitter sentiment analysis on a fixed set of Tweets retrieved all at once. Such studies are helpful in providing a snapshot of customers' opinions at a given time but offer little for airlines to rely on as a validated and ongoing reference. Seeing the shortcomings of existing studies and the airline industry's needs, we propose and build this pipeline to facilitate automated data retrieval, sentiment analysis, data visualization, and abnormality detection.

\section{Methods}
Our approach is composed of the following key steps:
\begin{itemize}
    \item Retrieving Data
    \item Natural Language Processing (NLP)
    \begin{itemize}
        \item Data Pre-processing
        \item Lexical Analysis
        \item Stemming and Lemmatization
        \item Vectorization
        \item Sentiment Analysis
    \end{itemize}
    \item Visualization through Time-Series Plots
    \item Anomaly Detection and Interpretation
\end{itemize}
Each of the above bullet points will be explained in the subsequent sections.

\subsection{Retrieving Data}
The first task is to acquire data from Twitter. An Application Programming Interface (API) allows communication between computer programs. Twitter's API, along with Python's built-in library Tweepy and proper access permission, makes it simple to access to Tweets archive with a simple API request. In this study, a total of ten major U.S. airlines: American, United, Southwest, Delta, JetBlue, Alaska, Allegiant Air, Frontier, Hawaiian, and Spirit are used for analysis.

We utilize Twitter's API v2 search endpoint \textit{GET /2/tweets/search/all}, which returns the complete history of public Tweets matching specific search requests. In order to construct an adequate search request, the following information needs to be specified: A \textit{ search query}, a starting/ending time in ISO-8601 format indicating the time range from which the Tweets will be provided, and the maximum results (currently capped at 500 due to Twitter's restriction). The response returned by the API is a list containing up to 500 Tweet Objects. 

A search query is a sequence of characters and operators accepted by the API search endpoints. Any Tweets that match the keywords in the query will be returned. The query also supports the use of Boolean operators such as AND, OR, and NOT. In the study, we used a combination of keywords with Boolean logic. For example, the search query used for United Airlines is:

$$@united \ OR \ united \ airlines \ lang:en \ -is-retweet$$

This query returns the Tweets that either contain mentions of United Airlines' official Twitter account (@united) or the keyword "united airlines." Important to mention, the query only returns Tweets in English and ignores Retweet. The search queries of the other nine airlines are built in the same fashion. Additionally, due to the restriction that only up to 500 results will be returned in a single request, for a single day's Tweets related to an airline, we use a script that divides a day into eight intervals and sends a request every three hours. By doing so, a maximum of 4000 Tweets could be collected for a single day, thus minimizing Tweets loss. We validate our approach by confirming that further increasing data retrieving frequencies does not increase the number of Tweets collected.

The retrieved Tweets are stored in a MySQL database hosted on the Google Cloud Platform. For every Tweet object, the following information is stored: the original Tweet's text, the Tweet's ID, the date this Tweet was created, author's ID, author's location (if any), the airline related to this Tweet, and the probabilities that this Tweet is positive/negative. The last two probabilities would be computed by the machine learning algorithms discussed in the following sections to categorize whether a Tweet is positive or negative.

\subsection{Natural Language Processing (NLP)}
As a branch of artificial intelligence (AI), NLP refers to the ability of a computer to understand and interpret human language, either in written or spoken form. In our study, our goal is to, with a proper sentiment analysis algorithm, develop an accurate NLP model that is able to determine whether a given Tweet is more likely to be associated with positive or negative sentiment.

Given a Tweet:\\
\centerline{\textit{@united United Airlines is the best airline! Thank you!}}
A human can almost immediately tell that this is a positive Tweet because words/phrases such as "best" and "thank you" naturally invoke a positive sentiment. However, for machines, pre-processing the input Tweets before feeding them to any machine learning model is crucial.

\subsubsection{Data Pre-processing}
Data pre-processing includes cleaning, normalization, transformation, feature extraction, and selection \cite{data1}. Proper data pre-processing can significantly improve the outcome of machine learning models because this process eliminates irrelevancy, redundancy, and other noises in the data. In our study, we focus on data cleaning, which includes lexical analysis and stemming; and data transformation using a vectorizer.

The initial step of data cleaning includes removing all other special characters besides alphabetical letters. In this step, we use a regular expression to scan each Tweet, removing punctuation, special characters, and emojis, which are part of the SNS culture. 
\subsubsection{Lexical Analysis}
Lexical analysis, also known as tokenization, is the process of separating word tokens from a sentence. Using the previous example Tweet, the following word tokens will be produced after lexing:

$$[united,\ united,\ airlines,\ is,\ the,\ best,\ airline,\ thank,\ you]$$
As mentioned above, words/phrases such as "best" and "thank you" provide a positive sentiment to this Tweet. While these words contribute to the overall sentiment of the Tweet, words such as "is" and "the" do not affect the over sentiment. Words like such are called stop words. During lexical analysis in NLP, stop words need to be removed from the word tokens in the original sentence so that the machine learning model does not use them as criteria to make predictions. Additionally, we remove airline keywords such as "united" and "airlines" so that we are left with the most relevant information related to the actual sentiment of the Tweet. The final tokens after removing stop words and other irrelevant keywords are:
$$[best,\ thank]$$

\subsubsection{Stemming and Lemmatization}
For grammatical reasons, Tweets use different forms of a word. For example, the word \textit{cancel} could appear as \textit{cancels}, \textit{canceled}, or \textit{canceling}. Similarly, they might also contain a word that belongs to the derivations of some other words. \textit{Bad} could appear as \textit{bad}, \textit{worse}, or \textit{worst}.

Stemming and lemmatization are standard methods used by search engines and chatbots to analyze the meaning behind a word. The former focuses on using the stem of the word, while the latter uses the context in which the word is being used \cite{lemma1}. Both algorithms are word normalization techniques in NLP in order to handle different variations of the same word properly \cite{lemma2}.

Stemming is a straightforward method that finds the stem of a word. When encountering different forms of the same word, stemming algorithms simply cut off the prefix or suffix and leaves a fragment of the word, usually the common part, as the stem. Table 1 shows some word stems generated by the Porter stemmer, a popular English stemming algorithm.

With word fragments, stemming results can be misleading and sometimes contain inaccuracies as to what the original words are. Thus, a more comprehensive approach, lemmatization, is often preferred over stemming. In this study, we used lemmatization to normalize the word tokens.

\begin{table}[!ht]
\caption{Comparison between Lemmatization and Stemming}\label{tab:versions}
    \begin{center}
      \begin{tabular}{l l l} 
     Word & Stem & Lemma\\\hline
     cities & citi & city\\
     flies & fli & fly\\ 
     flights & flight & flight\\
     services & servic & service\\
     terminals & termin & terminal\\\hline
     \end{tabular}   
    \end{center}
\end{table}

Unlike stemming, which cuts off part of the word and returns the stem, lemmatization reduces the word to its proper root form, known as the \textit{lemma}; the \textit{lemma} is also a word in the English dictionary. Additionally, lemmatization algorithms consider the part-of-speech of the given word, and different normalization rules will be applied for nouns, verbs, adjectives, and so on \cite{lemma2}.  With lemmatization algorithms, word tokens can be adequately normalized without loss of information.
Table 1 shows some frequent words from airline Tweets and why lemmatization is chosen over stemming.

\subsubsection{Vectorization}
Before feeding Tweets into the machine learning model, the final step of data pre-processing is vectorization. The goal of this step is to convert pre-processed tokens into vectors of real numbers using a vectorizer. Computers cannot take direct input of raw text but excels at handling numbers. The vectorizer we used is TF-IDF vectorizer, an abbreviation for \textit{Term Frequency Inverse Document Frequency}. The equation below illustrates how each value in the vector is computed:

$$w_{t,j} = tf(t, j) + log(\frac{N}{df(t)})$$

The essence behind this vectorizer is that it transforms a sample into a vector by calculating both the number of times that a word \textit{t} appears in a sample (that is, term frequency) and the number of samples that contain the word \textit{t} (that is, inverse document frequency). Each element of the transformed vector can be calculated using the above formula, where $tf(i, j)$ is the frequency of word $t$ in sample $j$; $df(t)$ is the number of samples containing word $t$; and $N$ is the total number of samples (in this study, total number of Tweets) in the training dataset.

\subsubsection{Sentiment Analysis}
The machine learning approach for predicting sentiments relies on ML algorithms such as probabilistic classifiers, linear classifiers, artificial neural networks (ANN), etc. These algorithms are based on mathematical and statistical approaches that find patterns within vectorized data \cite{sa1}. In this study, we did not use multi-layered neural networks but a linear classifier called the Support Vector Machines (SVM) as the sentiment prediction model, which can achieve just as high accuracy yet better run-time.

SVM is a mathematical entity. In essence, without delving into the complex equations, SVM classifies data by optimizing a cost function. The output is a line, or a hyperplane in higher dimensions, that separates the input data. Additionally, SVM chooses the separation hyperplane with the maximum margin \cite{svm1}.

\begin{figure}[!ht]
  \begin{center}
       \includegraphics[scale=0.5]{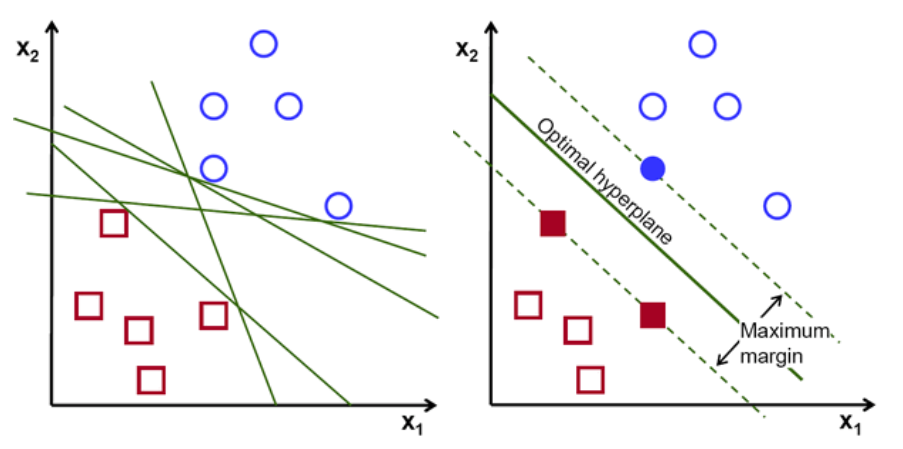}
  \end{center}
  \caption{SVM with different lines of separation (left) and the best one with maximum margin (right) \cite{svm2}}\label{fig:trial}
\end{figure}

Additionally, we provided the SVM with a kernel function. A kernel function of the form $$K(x, y) = <f(x), f(y)>$$ maps input data x and y from n-dimension to m-dimension by computing the dot product (m is usually much larger than n) \cite{kernel1}. When there is no clear linear separation in the data, mapping it to a higher dimension can result in linear separability and allow SVM to generate a higher dimensional hyperplane. The use of a kernel often results in better prediction accuracy. In our study, we use a radial basis function (RBF) kernel, or Gaussian kernel, computed using the following equation

$$K(x, y) = exp(-\frac{||x - y||^2}{2\sigma^2})$$

Equivalently, this kernel maps the input data to infinite dimensional space and thus often outputs the best prediction results.

The data used to train the SVM model is called "Twitter US Airline Sentiment" from Kaggle. This is the same dataset that is mentioned in the 1.2 Literature Review. It contains over ten thousand pre-labeled Tweets from February of 2015 related to popular U.S airlines. The dataset was first split into training data and testing data, and after pre-processing the training data and fitting them into the SVM model, we are able to obtain an SVM model with 91.86\% prediction accuracy on the testing data, meaning that, given an unknown Tweet, the model is able to determine whether a given Tweet is positive or negative about 91.86\% of the time. Figure 2 visualizes the confusion matrix, also known as the error matrix, when using the SVM model to predict the entire dataset.

\begin{figure}[!ht]
  \begin{center}
       \includegraphics[scale=0.5]{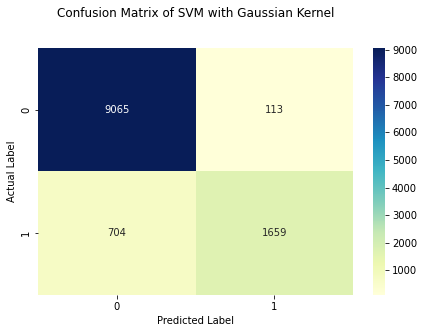}
  \end{center}
  \caption{Confusion Matrix Generated from the Training Data}\label{fig:trial}
\end{figure}

The output of each prediction contains two probabilities: the probabilities of positive and negative. In this binary classification study, where the labels only have positives or negatives, those probabilities are calibrated through Platt scaling \cite{svm3}.

\section{Results}
Using the data retrieval methods discussed in earlier sections, we collected over 1.3 million Tweets across ten airlines, with dates ranging from January 1 to July 12, 2022. After treating each Tweet with the  pre-processing methods discussed in previous sections, we pass these Tweets into the SVM model to predict the sentiment probabilities of being positive and negative. The greater probability of the two will be assigned as the predicted sentiment of the Tweet.

\subsection{Time Series Plots}
Time series plots will then be created to visualize the change over time in the number of positive and negative Tweets for a given airline. From figure 3, using American Airlines as an example, significantly more negative Tweets are posted than positive Tweets every day. And this is not uncommon for airlines as passengers are more likely to express their dissatisfaction or frustration on Twitter rather than offering compliments. 

\begin{figure}[!ht]
  \begin{center}
       \includegraphics[scale=0.5]{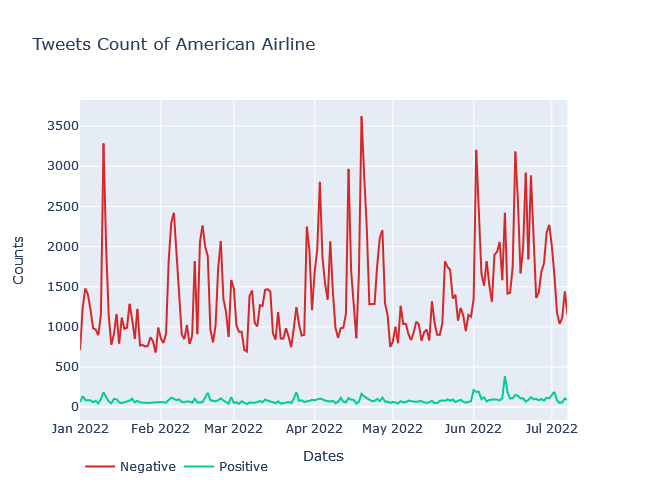}
  \end{center}
  \caption{Positive \& Negative Tweets counts of American Airlines from 1/1/2022 to 7/12/2022}\label{fig:trial}
\end{figure}

However, a mere count of negative and positive Tweets would not suffice for effective modeling customers' sentiments. While specific trends or anomalies, such as peaks of negative Tweets, could be observed from the plot, a more detailed and qualitative visualization is needed. To address this need, we  calculate the average sentiment over time and add normalized Bollinger Band to show the trend of sentiments.

A sentiment prediction made by the SVM model contains two probabilities. To calculate the average sentiment on a specific date, we sum up the difference between positive  and negative probability for each Tweet on that day. The average daily sentiment can take arbitrary values depending on the number of Tweets on that day. Through Z-normalization, the mean sentiment will be set to 0 and the standard deviation to 1.

A Bollinger Band is a technical analysis tool defined by a set of trend lines plotted two standard deviations away from a simple moving average (SMA), both positively and negatively \cite{bollinger1}. This type of plot is often used in the stock market to examine the falling and strengthening of assets. Since the lower and upper band constraint about 90\% of the data (being two standard deviations in both directions), any event that occurs above or below the band is considered a breakout. In our study, a breakout in the plot could potentially indicate disruptive events occurring in the airline industry that would cause an abrupt fall or rise in the average sentiment.

To construct the Bollinger band, we first calculate the simple moving average (SMA) of the Z-normalized average sentiment. In our study, the SMA indicates the average value of the average daily sentiment over a pre-set period \cite{bollinger2}. Figure 4 shows 7, 14, and 30 days SMA, along with the Z-normalized average, for the sentiment of American Airlines Tweets. Comparing this plot with Figure 2, a downfall in the SMA can often correspond to an increase in the number of negative Tweets. The obtrusive peaks of negatives between April-May and June-July both resulted in a drop in the SMA plot. During these periods, multiple breakouts occurred. In order to mathematically and visually identify these breakout points, we add Bollinger Bands above and below the SMA.

\begin{figure}[!ht]
  \begin{center}
       \includegraphics[scale=0.5]{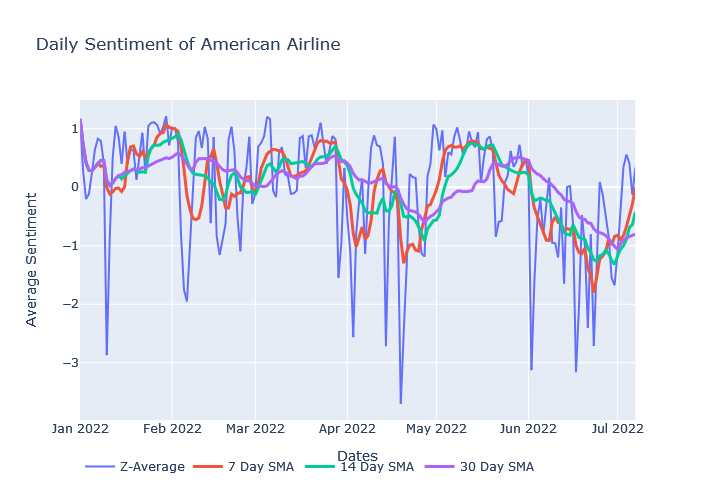}
  \end{center}
  \caption{Z-Normalized Average and SMAs}\label{fig:trial}
\end{figure}

\begin{figure}[!ht]
  \begin{center}
       \includegraphics[scale=0.5]{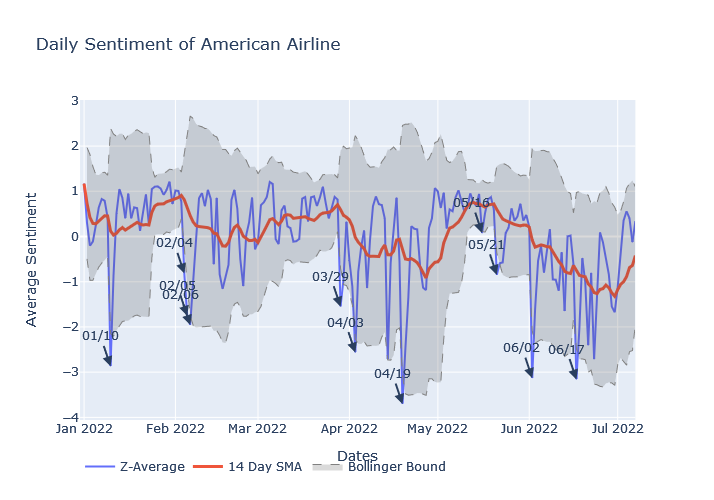}
  \end{center}
  \caption{Bollinger Bands with Breakout Points Labelled}\label{fig:trial}
\end{figure}

Figure 5 shows the 14-day SMA and its corresponding Bollinger Band, calculated by adding and subtracting two times the standard deviation of the SMA. This plot shows multiple breakout points when the daily sentiment average is below the Bollinger lower bound. As discussed above, breakout points signal potentially significant events or issues related to the airline industry. The following section will delve into word frequency analysis to provide a more cohesive insight into the research behind these breakout points and how the study could be applied to monitor customers' sentiments.

\subsection{Anomaly Detection and Word Frequency Analysis}
In order to understand the reason behind each breakout point, we perform word frequency analysis on the normalized word tokens. In this section, we will use multiple examples to demonstrate that Tweets' sentiment is indeed related to events in the airline industry, and we would be able to use time series plots to capture these abnormal sentiments.

Figure 5 shows multiple breakout points for American Airlines between February 2022 and March 2022. According to the Bureau of Transportation Statistics, American Airlines canceled about 8.27\% of its flights during this period, significantly over the average cancellation of 3.36\% in 2022. Using word frequency analysis, Table 2 shows the top five most common words that appear in the Tweets from the breakout points in February.

\begin{table}[!ht]
 \caption{Most Common Words from AA's Tweets in February's Breakout Points}\label{tab:versions}
\begin{center}
 \begin{tabular}{l l} 
 \hline
 Word & Frequency\\ \hline
 cancel & 1552\\
 hours & 1103\\ 
 call & 780 \\
 wait & 771 \\
 hold & 262\\
 \hline
 \end{tabular}
 \end{center}
\end{table}

An implication could be made that these breakpoints and falls in sentiment were mostly due to flight cancellations and, potentially, customer services (as the word \textit{call} and \textit{hold} also appeared quite frequently) and travelers were posting their complaints on Twitter. Similarly, using the same method, the breakout points on 1/10, 4/3, and those in May and June could all be explained by either flight cancellations or delays.

\begin{figure}[!ht]
  \begin{center}
       \includegraphics[scale=0.5]{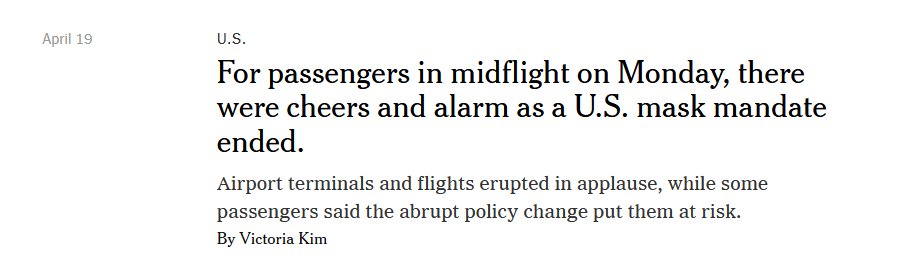}
  \end{center}
  \caption{Screen Shot of New York Times Article on Mask Mandate}\label{fig:trial}
\end{figure}
Besides flight cancellations, which are often one of the most common problems that travelers experience, other issues could manifest as anomalies in the plot. On 4/19, a breakout point occurred, and the daily sentiment reached the lowest in over six months. 4/19 was the day that mask mandates ended, and travelers were no longer required to put on their masks on planes. The end of the mask mandate happened so abruptly that many pilots announced it during mid-flight that passengers were free to take off their masks, according to a journal from New York Times \cite{news1}. The sudden end of the mask mandate caused concerns for some passengers, especially seniors or those flying with young kids. 

The last breakout point in Figure 5 is the one on 3/29. Using word frequency analysis, Table 3 might not immediately tell the issue that happened that day. However, after a simple search for the most common word \textit{sobapictures}, we found that this was referring to a Twitter user (\textit{@Sobapictures}) from Texas. On 3/29, he had his \$3,000 worth of luggage stolen on an American Airlines flight \cite{news2}. Later that day, he Tweeted:
"Last night I trusted \@AmericanAir
 w/ my equipment box and PAID \$60 for them to check in during my flight. After my arrival I opened my box to see \$3000 worth of equipment STOLEN." Soba had reached out to American Airline's customer services multiple times but did not receive a proper response. This news caused a significant wave of reactions on Twitter, and many users were enraged at the company's service. A snippet of a news report is shown in Figure 7.

\begin{table}[!ht]
 \caption{Most Common Words from AA's Tweets on 3/29}
\centering
\begin{center}
 \begin{tabular}{l l} 
 \hline
 Word & Frequency\\
 \hline
 sobapictures & 1166\\
 trust & 341\\ 
 check & 319 \\
 pay & 282 \\
 equipment & 273\\
 \hline
 \end{tabular}
 \end{center}
\end{table}

\begin{figure}[!ht]
  \begin{center}
       \includegraphics[scale=0.3]{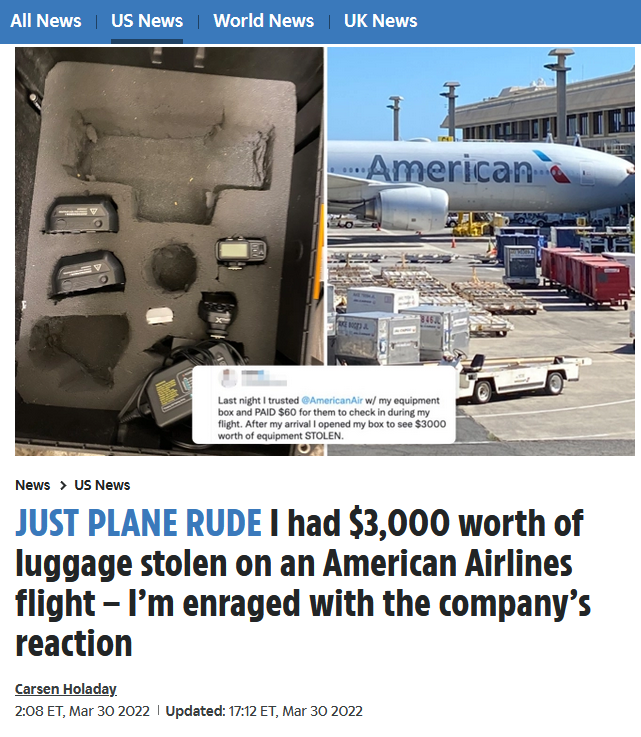}
  \end{center}
  \caption{Screen Shot from The Sun News}\label{fig:trial}
\end{figure}

\section{Conclusion}
This study retrieves and analyzes over 1.3 million Tweets posted by Twitter users related to ten major U.S. airlines from January ¬– July 2022. Through sentiment and lexical analysis, we are able to measure the public’s sentiments toward subject airlines with simple but effective metrics. Considering the nature of sentiment measures is time series, we apply popular financial concepts such as Bollinger Bands to visualize the trend of sentiment movement and help detect abnormalities in passenger sentiment. Using word frequency analysis, we provide context to help interpret abrupt sentiment movements. Using historical Tweets and reports, our approach is proven competent in detecting sharp temporal sentiment movements and providing convincing interpretations of such movements.     

Reliable and accessible data are critical for effective airline operations and management. Traditionally, airlines would rely on internal reporting mechanisms and external market studies to acquire information for status monitoring and process improvement. Data obtained through such channels are often outdated, unreliable, and expensive. Alternatively, User-Generated Contents (UGCs) related to commercial products and services are unsolicited, abundant, and genuine feedback from actual passengers. Such data, if properly utilized, would become a great addition to empower airlines’ emergency response, public communication, and customer relationship management.  

It needs to be noted that public sentiments based on Twitter data are subject to response bias. Generally speaking, users with stronger opinions, either positive or negative, are more likely to self-report their feedback. Therefore, when interpreting sentiment measures regarding airlines, readers should note that such data are not the outcome of random sampling but rather generated by a small group of passengers. However, this is not to suggest that UGCs are not valid information. Assuming responses from Twitter users are spontaneous and unsolicited, the sheer volume of Tweets each day is a good indicator to help airlines detect service disruptions. 

To facilitate potential users interested in our study, we have synthesized our analysis so far into a website that automatically fetches the newest Tweets. Users can have the freedom to navigate through figures presented in this paper, adjust the Bollinger Band, and search for different keywords to verify various events. Future research into analyzing Tweets can focus on developing a more sophisticated sentiment analysis model where neutral sentiment is also included along with positive and negative sentiments to create a more comprehensive categorization. Furthermore, while this study only utilizes Twitter data, other empirical studies may also be conducted on different SNS platforms for potential new insights.

\section{Author Contributions}
Shengyang Wu collected the data, built the analysis tool, performed the analysis, and drafted the paper. Yi Gao conceived and designed the study, performed time-series analysis, and drafted the paper. Both authors reviewed the results and approved the final version of the manuscript.

\newpage
\bibliographystyle{unsrt}  
\bibliography{references}

\end{document}